\documentclass[10pt, conference]{IEEEtran}
\IEEEoverridecommandlockouts

\usepackage{cite}
\usepackage{amsmath,amssymb,amsfonts}
\usepackage{algorithmic}
\usepackage{graphicx}
\usepackage{textcomp}
\usepackage{xcolor}
\def\BibTeX{{\rm B\kern-.05em{\sc i\kern-.025em b}\kern-.08em
    T\kern-.1667em\lower.7ex\hbox{E}\kern-.125emX}}
\usepackage{booktabs}
\usepackage{longtable}
\usepackage{tabularx}
\usepackage{hyperref}
\usepackage{multirow}
\usepackage[skip=2pt]{caption}
\usepackage{subcaption}
\usepackage{tcolorbox}
\usepackage{changepage}
\usepackage{pgfplots}
\usepackage{pgfplotstable}
\pgfplotsset{compat=1.18} 
\usepackage{array}
\hypersetup{hidelinks}

\begin{document}
\title{Mind2Drive: Predicting Driver Intentions from EEG in Real-world On-Road Driving}

\author{
Ghadah Alosaimi\textsuperscript{1,2},
Hanadi Alhamdan\textsuperscript{4,2},
Wenke E\textsuperscript{2},
Stamos Katsigiannis\textsuperscript{2},\\
Amir Atapour-Abarghouei\textsuperscript{2},
Toby P. Breckon\textsuperscript{2,3} \\[0.5em]
\textsuperscript{1}Department of Computer Science, Imam Mohammad Ibn Saud Islamic University, Saudi Arabia \\
Department of \{\textsuperscript{2}Computer Science, \textsuperscript{3}Engineering\}, Durham University, UK \\
\textsuperscript{4}Department of Computer Science, Princess Nourah bint Abdulrahman University, Saudi Arabia
}

\maketitle

\begin{abstract}
Predicting driver intention from neurophysiological signals offers a promising pathway for enhancing proactive safety in advanced driver assistance systems, yet remains challenging in real-world driving due to EEG signal non-stationarity and the complexity of cognitive–motor preparation. This study proposes and evaluates an EEG-based driver intention prediction framework using a synchronised multi-sensor platform integrated into a real electric vehicle. A real-world on-road dataset was collected across 32 driving sessions, and 12 deep learning architectures were evaluated under consistent experimental conditions. 
Among the evaluated architectures, TSCeption achieved the highest average accuracy (0.907) and Macro-F1 score (0.901). The proposed framework demonstrates strong temporal stability, maintaining robust decoding performance up to 1000 ms before manoeuvre execution with minimal degradation. Furthermore, additional analyses reveal that minimal EEG preprocessing outperforms  artefact-handling pipelines, and prediction performance peaks within a 400–600 ms interval, corresponding to a critical neural preparatory phase preceding driving manoeuvres. Overall, these findings support the feasibility of early and stable EEG-based driver intention decoding under real-world on-road conditions. Code: \url{https://github.com/galosaimi/Mind2Drive}.
\end{abstract}

\section{Introduction}
Accurate driver intention prediction is essential for road safety and effective interaction between human-driven vehicles and driver assistance or autonomous systems. Critical manoeuvres such as turning or braking are preceded by internal decision-making processes that are not directly observable from vehicle motion. Recent studies identify limitations in understanding human driving intentions as a major barrier to safe human–machine coexistence~\cite{xia2024driving}. Even modest gains in anticipation time, on the order of hundreds of milliseconds, can benefit collision avoidance and cooperative driving~\cite{liang2023eeg}.

Most driver intention prediction methods rely on external observations such as vehicle kinematics, trajectories, and traffic interactions. Machine learning and physics-informed models using speed, yaw rate, lane position, and inter-vehicle distances achieve high accuracy for manoeuvres such as lane changes and lane keeping~\cite{liu2023driving}. However, because these methods depend on observable post-manoeuvre motion cues, performance degrades with longer prediction horizons~\cite{liu2023driving, shi2025multi}.

Electroencephalography (EEG) offers a distinct sensing modality by directly capturing neural activity related to perception, decision-making, and motor preparation. Prior research has shown that intention-related neural signatures precede physical movement, which enables early motor intention decoding from EEG signals~\cite{idowu2021integrated}. Recent advances in deep learning further demonstrate that complex and fine-grained motor intentions can be decoded from EEG in real time, even under noisy conditions~\cite{ding2025eeg}.

In the driving domain, EEG-based intention prediction has been studied for tasks such as emergency braking, steering, and acceleration, demonstrating that EEG can anticipate driving intentions earlier than vehicle-based signals~\cite{liang2023eeg, li2024prediction}. However, most studies are conducted in simulators or controlled settings, and real-world on-road EEG experiments remain limited, with inconsistent preprocessing, labelling, and evaluation~\cite{tao2024multimodal}.

Another critical challenge lies in EEG preprocessing. EEG signals have inherently low signal-to-noise ratio and are strongly affected by motion artefacts and environmental interference in real-world driving. Comparative studies show that preprocessing choices substantially influence decoding performance, with no single pipeline being universally optimal~\cite{coelli2024selecting}. Nevertheless, preprocessing is often treated as a fixed design choice rather than as a systematic experimental factor.

Motivated by these gaps, this study investigates deep learning models, including transformer-based architectures, for EEG-based prediction of driver intentions during real-world on-road driving. Using synchronised EEG and high-precision vehicle kinematics, intentions are defined via motion-based labelling and predicted across multiple future horizons. In contrast to prior work, this study systematically evaluates EEG preprocessing pipelines and quantifies how early intentions can be decoded under realistic conditions, thereby advancing neuro-informed driver intention prediction for practical intelligent transportation systems. 

\section{Related Work}

We consider related work across the areas of vehicle-dynamics–based intention prediction (Section~\ref{subsec:rw-vehicle}), EEG motor-intention decoding (Section~\ref{subsec:rw-motorprep}), EEG-based driving intention prediction (Section~\ref{subsec:rw-eeg-driving}), human-centred driving systems using physiological cues (Section~\ref{subsec:rw-human-centered}), and methodological issues specific to real-world EEG (Section~\ref{subsec:rw-methods-realworld}).

\subsection{Driver Intention Prediction Using Vehicle Dynamics}\label{subsec:rw-vehicle} Prior studies infer driver intentions from vehicle-centric observations by modelling temporal patterns in kinematic and interaction variables. Time-sequenced and trajectory-based approaches, such as weighted hidden Markov models, have shown effective recognition of lane-change and lane-keeping behaviours from motion histories \cite{liu2023driving}. More recent physics-informed and interaction-aware learning frameworks further improve robustness and accuracy by incorporating vehicle dynamics and traffic constraints~\cite{shi2025multi}. However, because these methods infer intention from observable motion, their predictions remain closely coupled to kinematic onset, limiting characterisation of earlier, pre-movement intention stages.

\subsection{EEG-Based Intention and Motor Preparation}\label{subsec:rw-motorprep} Idowu et al. formulated motor intention recognition as a multi-class classification problem using fixed-length EEG segments aligned to explicitly cued upper-limb movements, achieving improved accuracy through deep architectures that combine spatial–spectral feature extraction with temporal modelling~\cite{idowu2021integrated}. This evaluation relies on segmented trials with known intention onset, assuming externally defined task timing. More recent work demonstrated real-time decoding of individual finger movements using end-to-end deep learning networks trained on densely supervised motor tasks, achieving low latency and high classification stability under laboratory conditions~\cite{ding2025eeg}. However, these approaches operate in tightly designed laboratory tasks in which road events, action timing, and movement boundaries are predefined, and subjects perform clearly separated, labelled movements in response to explicit cues (e.g., pressing a button, moving a finger, or lifting an arm). Consequently, they do not address the continuous, self-initiated intention formation or multi-horizon prediction required in real-world driving.

\subsection{EEG-Based Driver Intention Prediction}\label{subsec:rw-eeg-driving} Several studies have explored EEG-based intention prediction in driving, mainly focusing on braking, steering, or acceleration under controlled conditions. Liang et al. formulated emergency braking detection as a binary classification problem using EEG segments aligned to braking onset, demonstrating anticipation of braking actions in simulated driving~\cite{liang2023eeg}. Li et al. combined EEG and galvanic skin response signals to predict multiple driving intentions via multimodal fusion, while still relying on predefined manoeuvre events and simulator-based evaluation~\cite{li2024prediction}. More recent work applied few-shot transfer learning on neuromorphic hardware for individualised braking intention detection, but was limited to braking-only and structured tasks~\cite{lutes2025few}. Overall, these approaches focus on discrete, event-driven intentions and do not address continuous or multi-horizon prediction in real-world on-road driving.

\subsection{Human-Centred Driving Systems}\label{subsec:rw-human-centered} Recent work has incorporated human behavioural and neural signals as auxiliary inputs for autonomous driving, rather than directly decoding driver intention. Xia et al. integrated synchronised human behaviour data, extracted from EEG and other physiological signals, into end-to-end driving models in simulated environments, demonstrating that such signals can influence policy learning under controlled conditions~\cite{xia2024driving}. However, EEG was used as indirect supervision and evaluation was limited to simulation, without addressing explicit intention decoding or early prediction in real-world driving~\cite{duan2024enhancing}.

\subsection{Methodological Challenges and Real-World EEG}\label{subsec:rw-methods-realworld} A key challenge in EEG-based intention prediction studies is the robustness of signal preprocessing under real-world conditions. Coelli et al. showed that preprocessing choices can significantly affect decoding performance, with no single pipeline universally optimal across tasks and datasets~\cite{coelli2024selecting}. Tao et al. further highlighted that the limited availability of realistic large-scale driving datasets incorporating physiological signals constrains evaluation under realistic conditions~\cite{tao2024multimodal}.

Vehicle-centric predictors remain tethered to observable manoeuvres, while EEG-based methods largely stay confined to simulator/episodic paradigms, which leaves continuous intention decoding in real on-road settings underexplored.
\section{Methodology}
This section describes the experimental platform, data collection protocol, problem formulation, and deep learning models used in this study.

\begin{figure*}[!t]
    \centering
    \includegraphics[width=\textwidth]{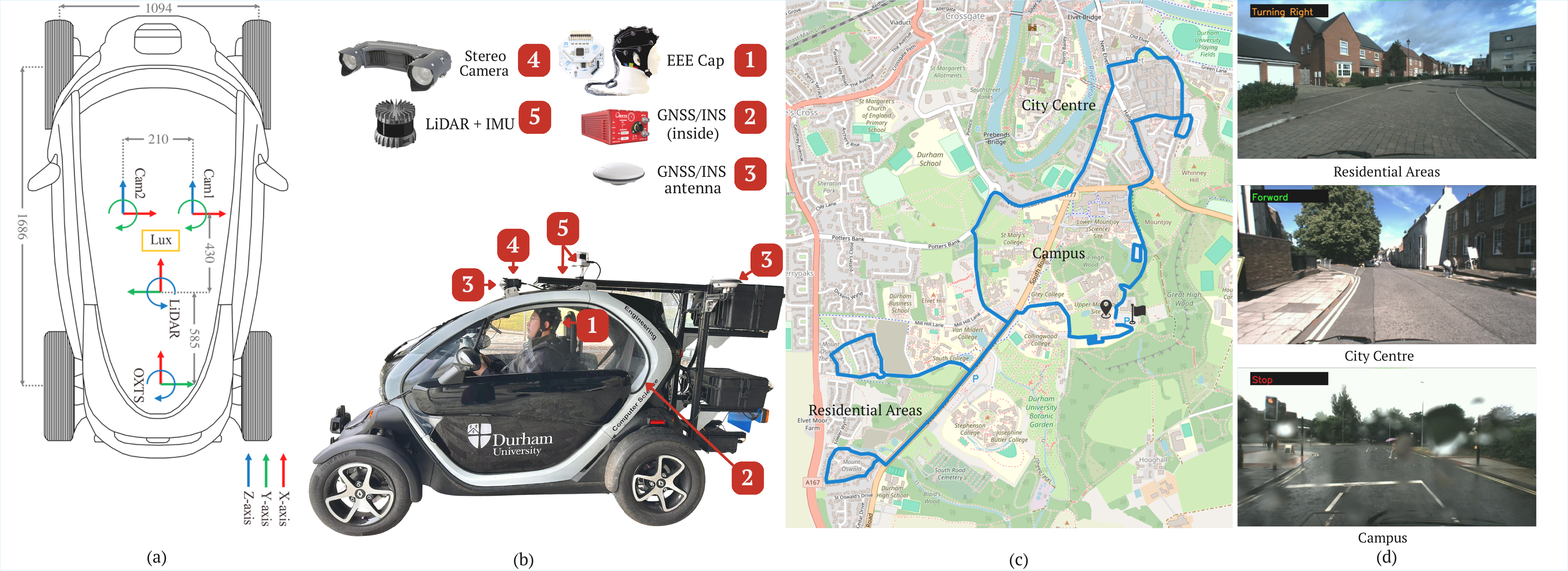}
    \caption{Experimental platform and data collection environment. (a, b) Instrumented Renault Twizy showing sensor placement and coordinate frames, including EEG, GNSS/INS, stereo cameras, and a LiDAR–IMU, with all sensor streams synchronised via ROS and recorded in rosbag format. (c) Predefined driving route across campus, city-centre, and residential areas. (d) Forward-facing camera views illustrating data collection under different environmental conditions.}
    \label{fig:env}
\end{figure*}

\begin{table*}[t]
    \centering
    \caption{Sensor specifications and data characteristics}
    \label{tab:sensors}
    \renewcommand{\arraystretch}{1.15}

    \begin{tabularx}{\textwidth}{l l c c X}
        \toprule
        \textbf{Sensor Type} & \textbf{Model} & \textbf{Sampling Rate} & \textbf{Channels / Beams} & \textbf{Measured Parameters} \\
        \midrule
        
        EEG & Gel-free OpenBCI cap (Cyton + Daisy) & 125 Hz & 16 &
        Scalp EEG (Fp1, Fp2, C3, C4, T3, T4, O1, O2, F7, F8, F3, F4, T5, T6, P3, P4) \\
        
        GPS / INS & OxTS RT3000v3 & 100 Hz & 2 antennas &
        Position (latitude, longitude), velocity $(v_x, v_y)$, yaw $(\psi)$, yaw-rate $(\dot{\psi})$ \\
        
        LiDAR + IMU & Ouster OS1-128 & 10 Hz & 128 &
        3D point clouds, angular velocity, linear acceleration \\
        
        Cameras & Carnegie Robotics Multisense S21 (stereo) & 30 Hz & 2 (left/right) &
        RGB images ($2048 \times 1088$ pixels) \\
        
        \bottomrule
    \end{tabularx}
\end{table*}

\subsection{Sensor Platform and Data Collection}
Data were collected using a Renault Twizy electric vehicle instrumented with a synchronised multi-sensor platform for real-world driving experiments. Vehicle layout and sensor placement are shown in \figurename~\ref{fig:env}, and sensor specifications are summarised in \tablename~\ref{tab:sensors}. All sensor streams were recorded using the Robot Operating System (ROS Noetic), providing unified timestamping for precise temporal alignment across data sources.

Vehicle motion was measured using a GNSS/INS unit with RTK correction via NTRIP, yielding accurate estimates of vehicle position, planar velocity, yaw angle, and yaw rate, which were used to derive ground-truth driving actions and intention labels. Additional sensors, including a stereo camera and LiDAR with an integrated IMU, were installed to provide visual and inertial context and support synchronisation. It should be noted that data from these sensors were not used as model inputs, but have been reserved for future work.

EEG signals were recorded using a wireless 16-channel OpenBCI Cyton and Daisy system with a gel-free electrode cap configured according to the international 10–20 system~\cite{openbci}. The cap provided coverage over frontal, central, temporal, parietal, and occipital regions, with electrode locations listed in \tablename~\ref{tab:sensors}.

Four participants (three male and one female, aged 25–45 years), each holding a valid UK driving licence, participated in the study. All had normal or corrected-to-normal vision and reported no history of neurological or psychiatric disorders. Written informed consent was obtained from all participants in accordance with institutional ethical guidelines.

Data collection took place on public roads in Durham, UK, covering campus, city-centre, and residential environments under varying weather conditions, including sunny, cloudy, and rainy. All participants followed the same predefined route of approximately 9 km to ensure comparable driving conditions while preserving realistic variability in traffic and driving behaviour. Representative environments and the driving route are shown in \figurename~\ref{fig:env}.

Each participant completed 8 driving sessions, yielding a total of 32 sessions. Each session involved a single traversal of the route with an average duration of approximately 22 minutes. Prior to data collection, participants completed two familiarisation sessions to acclimate to the vehicle, route, and experimental setup. Participants were instructed to drive naturally while complying with traffic regulations. All sessions meeting data-quality criteria, stable EEG recordings with electrode impedance below 10~k$\Omega$ and complete kinematic data without extended dropouts, were retained for analysis.

\subsection{Signal Synchronisation and Preprocessing}\label{subsec:m-prep}
EEG signals sampled at 125 Hz were used as the reference timeline for synchronisation within the ROS framework. Vehicle kinematic signals from the GNSS/INS system were linearly interpolated to the EEG timestamps, ensuring temporal alignment between neural activity and vehicle motion.

Prior to model input, EEG signals were normalised to ensure consistency across sessions and participants. No automated preprocessing pipeline was applied by default. To assess the impact of preprocessing, commonly used EEG cleaning strategies were subsequently evaluated, including PyPREP-based pipelines with band-pass filtering and optional RANSAC-based bad channel detection~\cite{bigdely2015prep}. Additional artifact handling approaches combining independent component analysis (ICA), artifact subspace reconstruction (ASR), and automated trial rejection were also tested~\cite{callan2024shredding, plechawska2023influence}. All configurations were applied consistently across models and sessions, and their effects are analyzed in Section~\ref{sec:prep_pipline_com}.

{\captionsetup[subfigure]{skip=3pt}
\begin{figure*}[t]
    \centering

    \begin{minipage}[t]{0.48\textwidth}
        \centering
        \includegraphics[width=\linewidth]{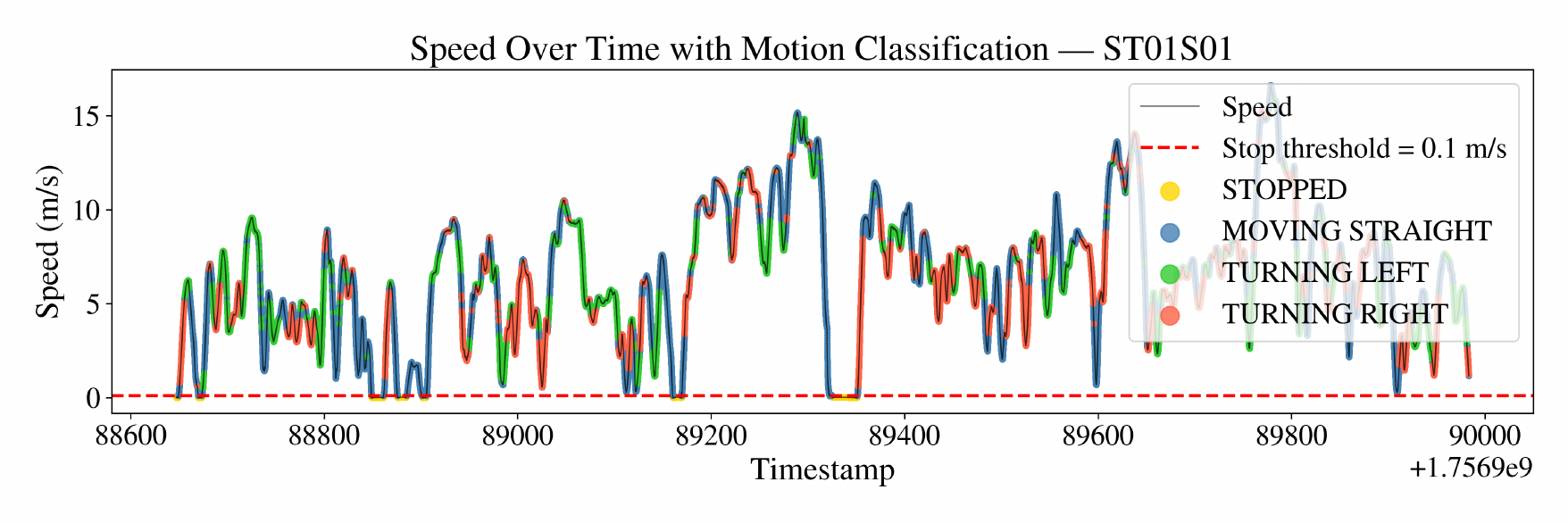}
        \subcaption{Vehicle speed over time with motion labels.}
        \label{fig:speed_time}
    \end{minipage}
    \hfill
    \begin{minipage}[t]{0.48\textwidth}
        \centering
        \includegraphics[width=\linewidth]{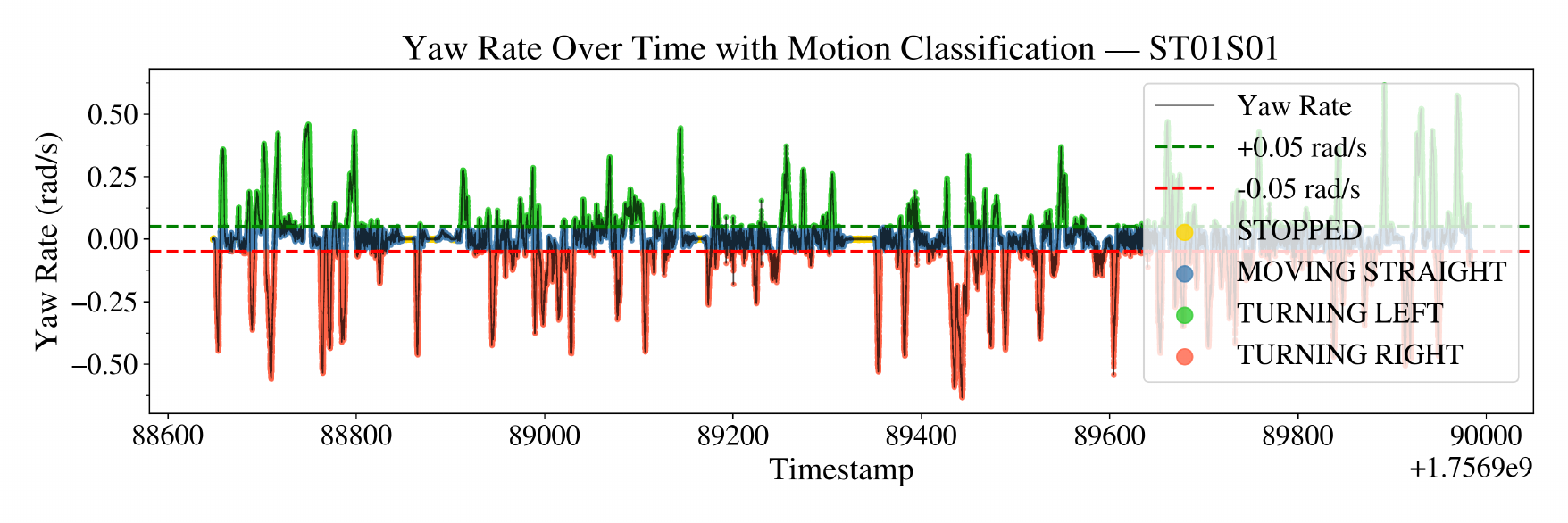}
        \subcaption{Yaw-rate over time with turning thresholds.}
        \label{fig:yaw_time}
    \end{minipage}

    \vspace{0.3em} 

    \begin{minipage}[t]{0.46\textwidth}
        \centering
        \includegraphics[width=\linewidth]{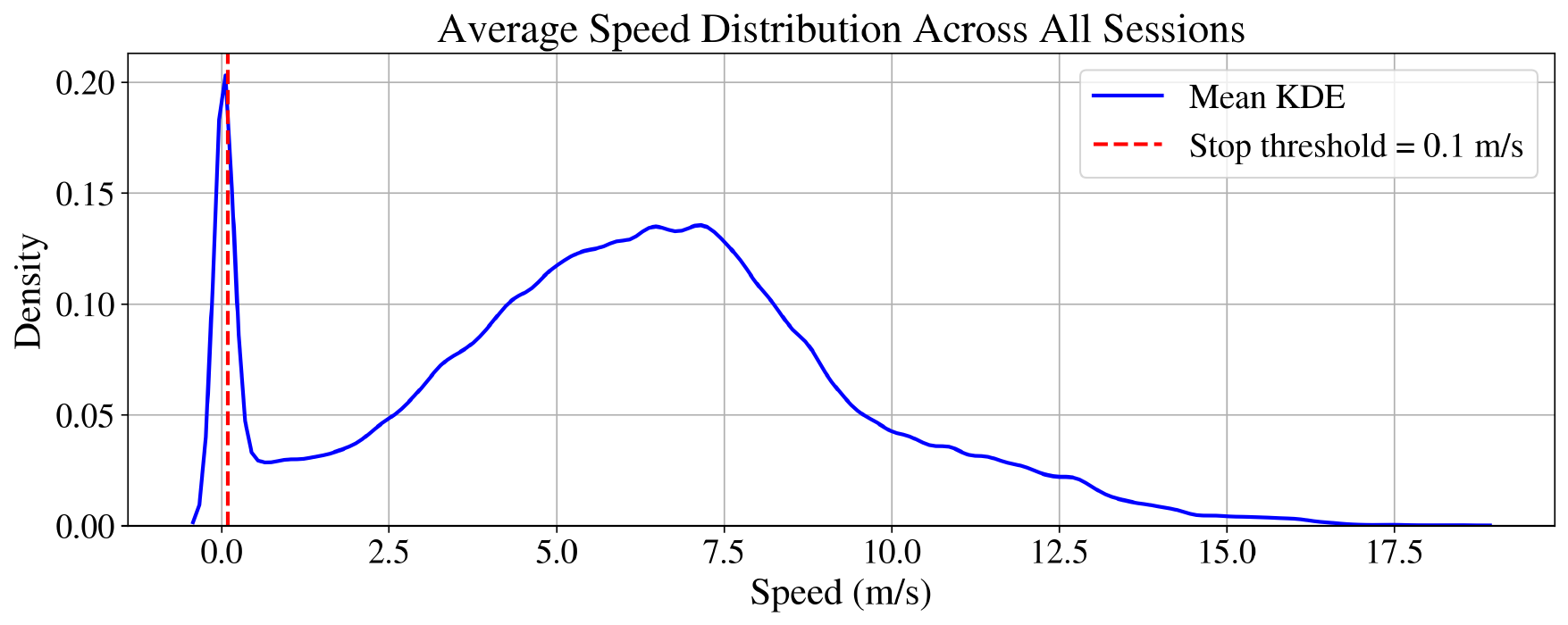}
        \subcaption{Kernel density estimate of vehicle speed.}
        \label{fig:speed_kde}
    \end{minipage}
    \hfill
    \begin{minipage}[t]{0.46\textwidth}
        \centering
        \includegraphics[width=\linewidth]{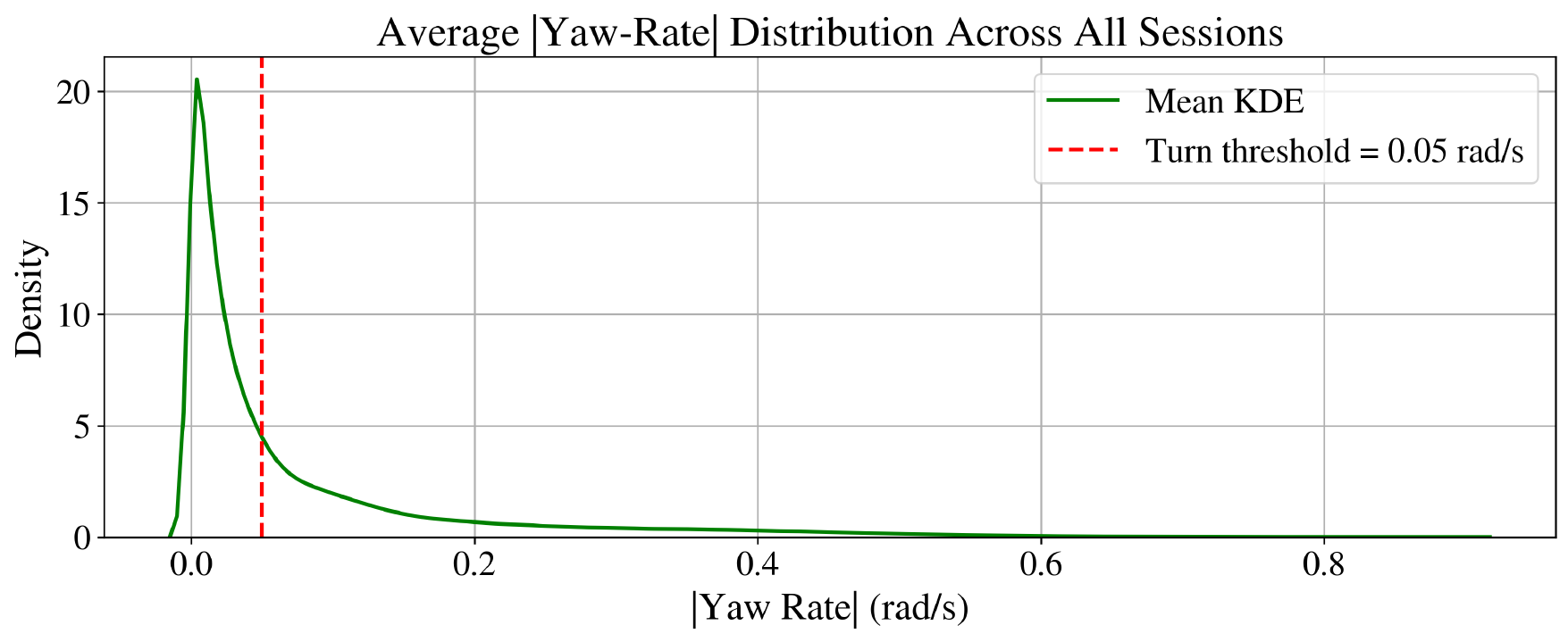}
        \subcaption{Kernel density estimate of yaw-rate magnitude.}
        \label{fig:yaw_kde}
    \end{minipage}

    \caption{Motion-based action labelling from vehicle kinematics. Representative time-series of speed and yaw rate with applied thresholds for a single session of a subject (top), and kernel density estimates across all sessions validating the selected stop and turning thresholds (bottom).}
    \label{fig:motion_labeling}
\end{figure*}
}

\subsection{Motion-Based Action Labelling}
Ground-truth driving actions were derived from vehicle kinematics measured by the GNSS/INS system. Continuous motion signals were converted into discrete manoeuvre labels based on vehicle speed and yaw-rate. Vehicle speed at time $t$ was computed from the planar velocity components as:
\begin{equation}
v(t) = \sqrt{v_x^2(t) + v_y^2(t)},
\end{equation}
where $v_x(t)$ and $v_y(t)$ denote the forward–backward and lateral velocity components in the road plane. Vehicle orientation was described by the yaw angle $\psi(t)$, and turning dynamics by the yaw rate $\dot{\psi}(t)$.

To distinguish forward and reverse motion, the direction of movement was computed as:
\begin{equation}
\theta_v(t) = \operatorname{atan2}\!\big(v_y(t),\, v_x(t)\big),
\end{equation}
and compared with the vehicle heading $\psi(t)$. The angular difference
\begin{equation}
\Delta \theta(t) = \operatorname{wrap}\!\big(\theta_v(t) - \psi(t)\big)
\end{equation}
was used to identify forward $(|\Delta \theta(t)| \le \pi/2)$ and  reverse motion $(|\Delta \theta(t)| > \pi/2)$.

Two thresholds were defined: a speed threshold $v_{\mathrm{th}}$ to separate stationary and moving states, and a yaw-rate threshold $\omega_{\mathrm{th}}$ to distinguish straight driving from turning. Vehicle motion at time $t$ was classified as:
\newline
\begin{adjustwidth}{0em}{0pt}
{\small
\textbf{Stopped:} \hspace*{3.3em} $v(t) < v_{\mathrm{th}}$ \\[2pt]
\textbf{Turning Left:} \hspace*{1.2em} $v(t) \ge v_{\mathrm{th}}$ AND $\dot{\psi}(t) > \omega_{\mathrm{th}}$ \\[2pt]
\textbf{Turning Right:} \hspace*{0.5em} $v(t) \ge v_{\mathrm{th}}$ AND $\dot{\psi}(t) < -\omega_{\mathrm{th}}$ \\[2pt]
\textbf{Moving Straight:} $v(t) \ge v_{\mathrm{th}}$ AND $|\dot{\psi}(t)| \le \omega_{\mathrm{th}}$ AND $|\Delta \theta(t)| \le \frac{\pi}{2}$ \\[2pt]
\textbf{Reverse:} \hspace*{3.4em} $v(t) \ge v_{\mathrm{th}}$ AND $|\Delta \theta(t)| > \frac{\pi}{2}$
}
\end{adjustwidth} 
\hfill \break
Thresholds were selected through visual inspection of representative time-series and validated using global distributions across all sessions. As shown in \figurename~\ref{fig:motion_labeling}, speed and yaw-rate signals exhibit clear separation between stationary, straight-driving, and turning regimes, which is further confirmed by kernel density estimates. The resulting vehicle motion labels were additionally validated through manual visual inspection of the recorded videos for a subset of the samples. Reverse and stopped states were excluded from the final action set, as reverse manoeuvres were not part of the predefined route and the study focuses on continuous forward-driving behaviour.

\subsection{Intention Prediction Problem Formulation}
This study aims to predict a driver’s future driving intention from EEG signals prior to manoeuvre execution. The problem is formulated as a supervised task in which neural activity preceding a manoeuvre is used to predict a future action.

Let $X(t) \in \mathbb{R}^{C \times T}$ denote the multichannel EEG signal observed up to time $t$, where $C$ is the number of EEG channels and $T$ is the number of temporal samples in the observation window. Let $y(t)$ denote the discrete driving action being executed by the vehicle at time $t$, obtained from motion-based labelling. A future-intention label at prediction horizon $\Delta$ is defined as
\begin{equation}
y_{\Delta}(t) = y(t + \Delta),
\end{equation}
with $\Delta \in \{100, 200, 300, \ldots, 1000\}$~ms. The learning objective is to map an EEG window $X(t-w:t)$ to the corresponding future action $y_{\Delta}(t)$ , where $w$ denotes the window length. The prediction task is restricted to the action set: $y \in \{\text{Forward},\ \text{Turn--Left},\ \text{Turn--Right}\}$.

Multiple prediction horizons are considered to quantify how early driver intentions can be decoded from EEG. Short horizons ($100$--$300$~ms) correspond to near-term motor preparation, while longer horizons ($500$--$1000$~ms) reflect earlier stages of cognitive intention formation.

\subsection{Experimental Design and Evaluation Protocol}\label{subsec:m-exp}
Twelve deep learning architectures were evaluated to provide a comprehensive assessment of EEG-based driver intention classification across diverse model families. The selected models span convolutional, recurrent, and attention-based paradigms. Convolutional models extract local spatiotemporal EEG patterns through temporal and spatial filtering~\cite{iwama2024eeg}, and include TSCeption~\cite{ding2022tsception}, ShallowConvNet and DeepConvNet~\cite{schirrmeister2017deep}, EEGNet~\cite{lawhern2018eegnet}, CCNN~\cite{yang2018continuous}, DGCNN~\cite{song2018eeg}, CNN1D~\cite{taghizadeh2024eeg}, and STNet~\cite{zhang2022ganser}. Recurrent models capture sequential EEG dependencies~\cite{sikka2020investigating} and are represented by LSTM and GRU~\cite{zhang2021deep}. Attention-based architectures combine convolutional feature extraction with mechanisms for modelling long-range temporal relationships~\cite{ma2024attention}, and include EEGConformer~\cite{song2022eeg} and ViT~\cite{dosovitskiy2020image}. This selection was designed to cover a broad spectrum of architectural inductive biases, temporal modelling capabilities, and spatial feature extraction strategies commonly adopted in EEG decoding research.

Furthermore, based on preliminary experimentation and the prohibitively large number of possible configuration combinations, four representative models, EEGNet, ShallowConvNet, GRU, and EEGConformer, were selected for an ablation study, as they reflect complementary architectural families. The ablation study focused on identifying optimal data preparation strategies, EEG preprocessing pipelines, and the impact of using 8 versus 16 EEG channels.

All models were trained and evaluated using identical data splits and windowing configurations to ensure that performance differences reflect preprocessing or architectural effects rather than data preparation variability. To ensure fair comparison, standardised training settings were used across models. A batch size of 128 and a maximum of 2,000 training epochs were applied, with the Adam optimiser; all models used a learning rate of 0.001 except EEGConformer (1e-4), with all learning rates remaining in the same order of magnitude as those reported in their original papers. Loss functions were selected according to model output formulations, with class-weighted variants used where necessary to address class imbalance. The model checkpoint achieving the best Macro-F1 score during held-out validation was selected for final evaluation. Data splitting and training were performed at the session level to preserve temporal continuity and avoid information leakage. All experiments were conducted on multiple NVIDIA RTX A6000 GPUs to ensure consistent computational conditions.

Performance was primarily assessed using Macro-F1 to account for class imbalance between forward-driving and turning manoeuvres. Balanced accuracy and per-class recall were also reported to provide complementary performance insights.

\begin{figure}[t]
\centering
\includegraphics[width=\linewidth]{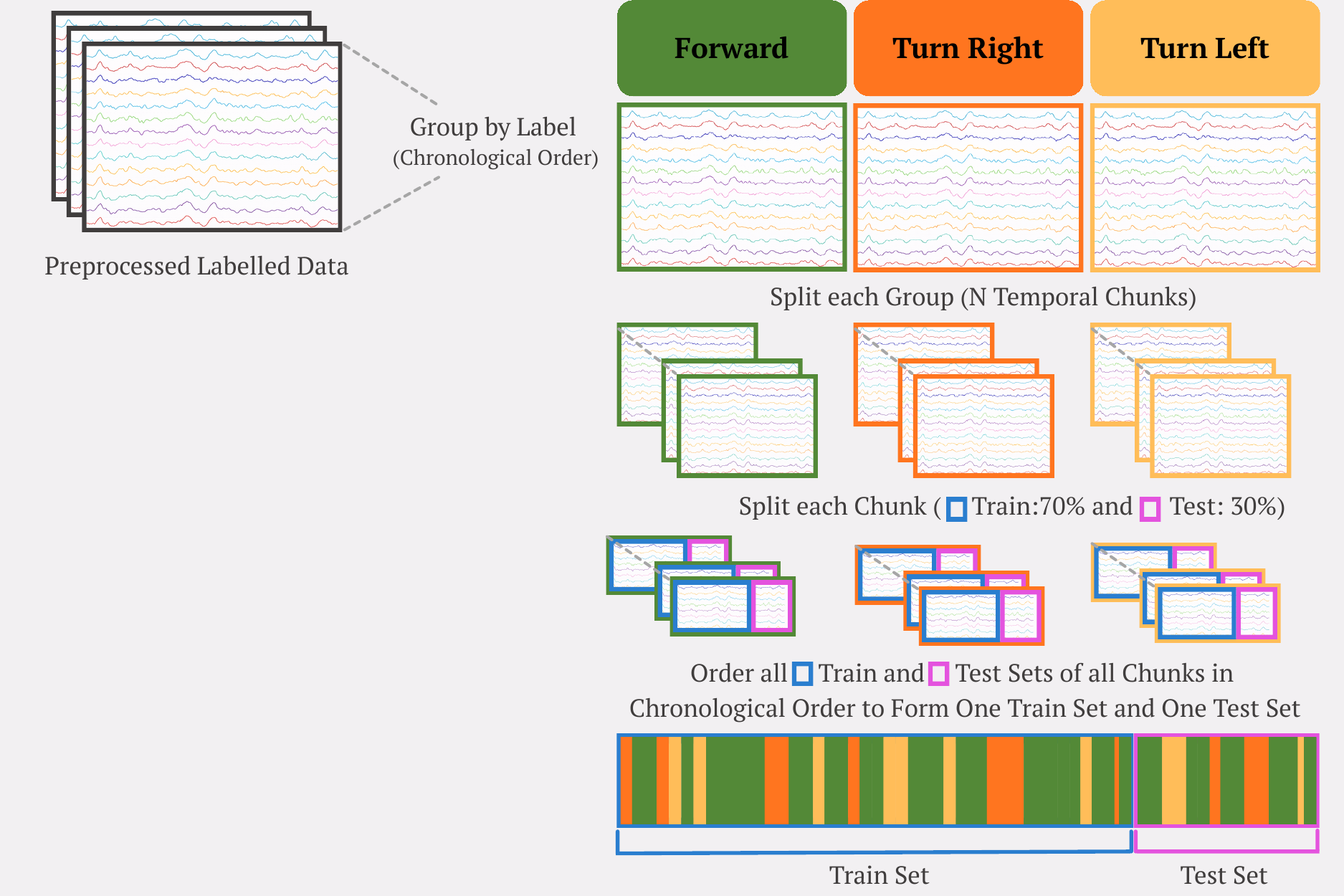}
\caption{EEG recording is grouped by class and ordered chronologically. Each group is split into 70\%-train and 30\%-test segments using temporal chunks, then re-ordered to form the final train and test sets.}
\label{fig:data_construction}
\end{figure}

\section{Experimental Results}
This section presents a systematic evaluation of the proposed EEG-based driver intention prediction framework, assessing the effects of data preparation strategies, preprocessing pipelines, and model architectures.

\subsection{Evaluation of data preparation strategies}
The impact of data preparation choices on EEG-based driver intention prediction performance is analyzed prior to preprocessing and model-specific analysis, with the goal of identifying a robust strategy for subsequent experiments.

A systematic ablation was conducted over data splitting, windowing, and label aggregation strategies. To prevent temporal leakage, data were split on continuous and time-ordered EEG signals before windowing. Windowing was applied independently within training and test sets, and oversampling, when used, was restricted to the training data. Twelve windowing configurations were evaluated by combining four window lengths (0.5s, 1.0s, 2.0s, and 5.0s) with two step sizes (non-overlapping and 50\% overlap). Window-level labels were assigned using either a majority rule, which assigns the most frequent action within a window, or a rejection rule, which retains only windows containing a single, consistent manoeuvre and discards windows spanning action transitions, thereby ensuring manoeuvre purity. The rejection rule was applied only to 0.5s and 1.0s windows, as longer windows showed excessive rejection rates. Each configuration was tested using two data-splitting strategies: a standard temporal split (70\%/30\%) and a label-stratified split preserving class balance while maintaining temporal order, as illustrated in (\figurename~\ref{fig:data_construction}). Three imbalance-handling settings were also considered: no oversampling, random oversampling, and TimeGAN-based augmentation \cite{yoon2019time}.

Performance was evaluated using average Macro-F1 and balanced accuracy across the four models selected for the ablations to reduce architecture-specific bias, with only top configurations reported. Across all settings, a 1.0s window with 50\% overlap and the rejection labelling rule achieved the best performance, outperforming majority voting. The choice of data split had a substantial impact on performance. The label-stratified split significantly outperformed the standard split, achieving an average accuracy of 73.88\% compared to 36.53\%, corresponding to an improvement of approximately 37\%. This highlights the importance of preserving class balance between training and test sets for reliable intention prediction. On average, 27.1\% of windows per session were rejected due to ambiguous labels, reducing the mean number of windows from approximately 5,096 to 3,731 per session. Despite this reduction, rejecting ambiguous windows improved label consistency and downstream classification performance, with the rejection rule outperforming majority voting by approximately 11\%.

The effect of class imbalance handling is summarised in \tablename~\ref{tab:combined_results}. Random oversampling achieved the highest average Macro-F1 (0.873) across four models, outperforming both the baseline configuration without oversampling (0.828) and TimeGAN-based augmentation (0.811), indicating improved performance on minority driving intention classes. While the baseline configuration achieved the highest average balanced accuracy (0.858), its Macro-F1 was lower than that of random oversampling. TimeGAN achieved competitive balanced accuracy (0.852), but did not consistently outperform random oversampling, suggesting limited benefit for this task.

Based on this ablation, a label-stratified split with random oversampling, combined with 1.0s windows, 50\% overlap, and the rejection labelling rule, was selected as the fixed data preparation strategy for all subsequent experiments.

\begin{table}[t]
\centering
\caption{Comparison of the top-performing data preparation strategies, selected from all evaluated combinations, reported using the average metrics across the four models used for ablations. }
\label{tab:combined_results}

\footnotesize
\begin{tabularx}{\linewidth}{@{}c c c X c c@{}}
\toprule
\textbf{Win.} &
\textbf{Ov.} &
\textbf{Agg.} &
\textbf{Sampling Strategy} &
\textbf{Macro-F1} &
\textbf{Bal. Acc} \\
\midrule

\textbf{1.0s} & \textbf{50\%} & \textbf{Reject} & \textbf{Random} 
& \textbf{0.873 $\pm$ 0.038} & 0.838 $\pm$ 0.044 \\

1.0s & 0\% & Reject & Random 
& 0.855 $\pm$ 0.042 & 0.812 $\pm$ 0.044 \\

1.0s & 50\% & Reject & None 
& 0.828 $\pm$ 0.059 & 0.858 $\pm$ 0.037 \\

1.0s & 50\% & Reject & TimeGAN 
& 0.811 $\pm$ 0.048 & 0.852 $\pm$ 0.027 \\

0.5s & 25\% & Reject & Random 
& 0.806 $\pm$ 0.049 & 0.780 $\pm$ 0.056 \\

\bottomrule
\end{tabularx}

\vspace{2pt}
\begin{flushleft}
\footnotesize\emph{Note:} Win. = window length; Ov. = overlap; Agg. = aggregation; Bal. Acc = Balanced Accuracy; None indicates no sampling (baseline).
\end{flushleft}

\end{table}

\subsection{Evaluation of EEG Preprocessing Pipelines}
\label{sec:prep_pipline_com}

The effect of commonly used EEG preprocessing pipelines, described in Section~\ref{subsec:m-prep}, on model performance is evaluated using the above fixed data preparation strategy, with the aim of assessing whether advanced preprocessing offers benefits over minimal preprocessing.
Results in \tablename~\ref{tab:preprocessing_comparison} show a consistent trend across all models: minimal preprocessing achieved the best overall performance. The baseline configuration using only z-score normalisation (without PyPREP) yielded the highest average accuracy (0.839) and Macro-F1 (0.873), indicating that raw EEG signals already contain discriminative information for driver intention prediction. Applying PyPREP without RANSAC produced comparable results but did not improve performance. In contrast, pipelines incorporating aggressive artefact handling, such as RANSAC-based channel rejection, ICA, ASR, and automated trial rejection consistently degraded performance, suggesting that intention-related neural information may be removed alongside noise in real-world EEG.

Based on these results, the baseline preprocessing configuration without PyPREP was selected for all subsequent experiments to maximise performance while maintaining simplicity and reproducibility.

\begin{table}[t]
\centering
\caption{Comparison of different EEG preprocessing pipelines, reported using the average metrics across the four models used for ablations.}
\label{tab:preprocessing_comparison}

\begin{tabular*}{\columnwidth}{@{\extracolsep{\fill}}lcc}
\toprule
\textbf{Preprocessing Pipeline} & \textbf{Macro-F1} & \textbf{Bal. Acc} \\
\midrule
\textbf{No PyPREP (Baseline)}   & \textbf{0.873 $\pm$ 0.038} & \textbf{0.839 $\pm$ 0.047} \\
PyPREP (No RANSAC)              & 0.823 $\pm$ 0.062          & 0.831 $\pm$ 0.060 \\
AutoReject Only                 & 0.582 $\pm$ 0.143          & 0.700 $\pm$ 0.099 \\
PyPREP (RANSAC)                 & 0.557 $\pm$ 0.084          & 0.608 $\pm$ 0.088 \\
Raw Filter Only                 & 0.548 $\pm$ 0.101          & 0.599 $\pm$ 0.109 \\
ICA Only                        & 0.545 $\pm$ 0.101          & 0.594 $\pm$ 0.115 \\
ICA + ASR                       & 0.527 $\pm$ 0.113          & 0.585 $\pm$ 0.147 \\
ASR Only                        & 0.524 $\pm$ 0.120          & 0.570 $\pm$ 0.161 \\
ICA + AutoReject                & 0.523 $\pm$ 0.136          & 0.614 $\pm$ 0.157 \\
ASR + AutoReject                & 0.508 $\pm$ 0.149          & 0.624 $\pm$ 0.148 \\
ICA + ASR + AutoReject          & 0.504 $\pm$ 0.123          & 0.583 $\pm$ 0.148 \\
\bottomrule
\end{tabular*}
\end{table}

\begin{table}[t]
\centering
\caption{Performance of 8- and 16-channel EEG configurations}
\label{tab:channel_comparison}
\renewcommand{\arraystretch}{1.15}
\setlength{\tabcolsep}{2.9pt}

\begin{tabular*}{\columnwidth}{@{\extracolsep{\fill}}lcc|cc}
\toprule
\textbf{Model} 
& \multicolumn{2}{c}{\textbf{8 Channels}} 
& \multicolumn{2}{c}{\textbf{16 Channels}} \\
\cmidrule(lr){2-3} \cmidrule(lr){4-5}
& \textbf{Macro-F1} & \textbf{Bal. Acc} 
& \textbf{Macro-F1} & \textbf{Bal. Acc} \\
\midrule
EEGConformer   & 0.837 & 0.857 & 0.901 & 0.878 \\
EEGNet         & 0.636 & 0.662 & 0.820 & 0.777 \\
GRU            & 0.759 & 0.782 & 0.870 & 0.840 \\
ShallowConvNet & 0.833 & 0.835 & 0.902 & 0.860 \\
\bottomrule
\end{tabular*}
\vspace{2pt}
\begin{flushleft}
\vspace{-4pt}
\footnotesize\emph{Note:} Standard deviations across all entries range from 2.7\% to 6.8\%.
\end{flushleft}
\end{table}

\subsection{Effect of Spatial EEG Resolution (8 vs. 16 Channels)}
To evaluate the impact of spatial resolution, models were trained using both 8-channel and 16-channel EEG configurations under identical data preparation and preprocessing settings (\tablename~\ref{tab:channel_comparison}). For the 8-channel configuration, electrodes F3, F4, F7, F8, C3, C4, P3, and P4 were selected based on established neurophysiological evidence linking frontal regions to executive control and response inhibition, central regions to motor preparation, and parietal regions to spatial attention and sensorimotor integration~\cite{khaliliardali2015action, haufe2011eeg}. Across all architectures, increasing the number of channels from 8 to 16 improved accuracy and Macro-F1, indicating that additional spatial information enhances intention-related EEG representations. Although the 8-channel configuration produced competitive results, particularly for convolutional models, the 16-channel setup consistently achieved better generalisation and class discrimination. Consequently, the 16-channel configuration is used for all subsequent experiments.

\subsection{Evaluation of Model Performance}
This section presents a systematic comparison of twelve aforementioned deep learning architectures (Section~\ref{subsec:m-exp}) for EEG-based driver intention classification. All models were trained using the previously selected configuration. 

\subsubsection{Overall Performance Across Prediction Horizons}
\tablename~\ref{tab:model_comparison} summarises model performance averaged across all prediction horizons (0--1000 ms). The top-performing models (TSCeption, CNN1D, EEGConformer and ShallowConvNet) achieved the highest Macro-F1 scores, with $\text{F1} \geq 0.870$, and demonstrated robust performance across prediction horizons. Among them, TSCeption attained the best overall results ($\text{F1} = 0.901 \pm 0.003$), followed by CNN1D and EEGConformer. Notably, TSCeption exhibited exceptional stability across all horizons, indicating strong robustness to temporal offsets from action onset. Several models (DGCNN, GRU, STNet, CCNN, LSTM, and EEGNet) achieved moderate performance ($0.762 \leq \text{F1} \leq 0.834$). While these models provided reliable decoding, they were consistently outperformed by architectures designed for multi-scale temporal–spectral modelling. In contrast, DeepConvNet and ViT exhibited substantially lower Macro-F1 scores. Notably, while ViT achieved a moderate average classification accuracy of ($0.588$), its low Macro-F1 score ($0.572$)  indicates poor recognition of minority intention classes, suggesting limitations in handling class imbalance despite reasonable overall accuracy.

\begin{table}[t]
\centering
\caption{Model performance averaged across all prediction horizons}
\label{tab:model_comparison}

\begin{tabularx}{\columnwidth}{c X c c}
\toprule
\textbf{Ref.} & \textbf{Model} & \textbf{Macro-F1} & \textbf{Bal. Acc} \\ \midrule
~\cite{ding2022tsception} & TSCeption & \textbf{0.901 $\pm$ 0.003} & \textbf{0.907 $\pm$ 0.003} \\
~\cite{taghizadeh2024eeg} & CNN1D & 0.894 $\pm$ 0.002 & 0.899 $\pm$ 0.002 \\
~\cite{song2022eeg} & EEGConformer & 0.874 $\pm$ 0.004 & 0.887 $\pm$ 0.004 \\
~\cite{schirrmeister2017deep} & ShallowConvNet & 0.870 $\pm$ 0.005 & 0.869 $\pm$ 0.005 \\
~\cite{zhang2021deep} & LSTM & 0.834 $\pm$ 0.008 & 0.843 $\pm$ 0.012 \\
~\cite{zhang2021deep} & GRU & 0.830 $\pm$ 0.005 & 0.841 $\pm$ 0.004 \\
~\cite{song2018eeg} & DGCNN & 0.814 $\pm$ 0.004 & 0.799 $\pm$ 0.009 \\
~\cite{zhang2022ganser} & STNet & 0.805 $\pm$ 0.003 & 0.814 $\pm$ 0.006 \\
~\cite{yang2018continuous} & CCNN & 0.803 $\pm$ 0.034 & 0.813 $\pm$ 0.045 \\
~\cite{lawhern2018eegnet} & EEGNet & 0.762 $\pm$ 0.004 & 0.780 $\pm$ 0.004 \\
~\cite{schirrmeister2017deep} & DeepConvNet & 0.641 $\pm$ 0.012 & 0.472 $\pm$ 0.017 \\
~\cite{dosovitskiy2020image} & ViT & 0.572 $\pm$ 0.007 & 0.588 $\pm$ 0.019 \\
\bottomrule
\end{tabularx}
\end{table}

\subsubsection{Baseline and Optimal Horizon Performance}
To further characterise model behaviour, \tablename~\ref{tab:model_performance} reports performance at the baseline (0 ms) and at each model’s optimal prediction horizon. TSCeption achieved the strongest baseline performance ($\text{F1}=0.899$) and peaked at a 400 ms horizon. CNN1D, EEGConformer and ShallowConvNet followed closely, showing robust performance with modest gains at future horizons. EEGConformer also achieved competitive results, with ($\text{F1}=0.881$) peaking at 600 ms, highlighting the benefit of attention mechanisms for modelling temporally distributed EEG patterns.

Recurrent architectures (GRU and LSTM) and EEGNet achieved moderate performance, with consistent but limited improvements at future horizons, typically peaking around 400-500 ms. In contrast, general-purpose architectures such as ViT and DeepConvNet showed poor generalisation, underscoring the importance of domain-specific inductive biases for EEG-based intention prediction.

\begin{table}[t]
\centering
\caption{Model performance at baseline (0ms) and optimal prediction horizon}
\label{tab:model_performance}
\renewcommand{\arraystretch}{1.1}
\setlength{\tabcolsep}{4pt}

\begin{tabular*}{\columnwidth}{@{\extracolsep{\fill}}lccccc}
\toprule
\multirow{2}{*}{\textbf{Model}} 
& \multicolumn{2}{c}{\textbf{Baseline (0ms)}} 
& \multicolumn{3}{c}{\textbf{Optimal Horizon}} 
\\
\cmidrule(lr){2-3} \cmidrule(lr){4-6}
& \textbf{Macro-F1} & \textbf{Bal. Acc.} & \textbf{Macro-F1} & \textbf{Bal. Acc.} & \textbf{Time}\\
\midrule
TSCeption       & 0.899 & 0.905 & 0.905 & 0.914 & 400ms \\
CNN1D           & 0.894 & 0.899 & 0.899 & 0.905 & 400ms \\
EEGConformer    & 0.870 & 0.882 & 0.881 & 0.891 & 600ms \\
ShallowConvNet  & 0.868 & 0.868 & 0.877 & 0.876 & 400ms \\
LSTM            & 0.835 & 0.843 & 0.843 & 0.857 & 500ms \\
GRU             & 0.835 & 0.846 & 0.837 & 0.848 & 400ms \\
DGCNN           & 0.811 & 0.795 & 0.820 & 0.810 & 600ms \\
STNet           & 0.802 & 0.808 & 0.811 & 0.815 & 500ms \\
CCNN            & 0.755 & 0.771 & 0.849 & 0.856 & 1000ms \\
EEGNet          & 0.761 & 0.778 & 0.769 & 0.786 & 400ms \\
DeepConvNet     & 0.642 & 0.479 & 0.666 & 0.496 & 900ms \\
ViT             & 0.575 & 0.601 & 0.585 & 0.615 & 800ms \\
\bottomrule
\multicolumn{6}{l}{\footnotesize\emph{Note:} Standard deviations across all entries range from $\pm$0.03 to 0.23.}
\end{tabular*}
\end{table}

\subsection{Temporal Prediction and Horizon Analysis}
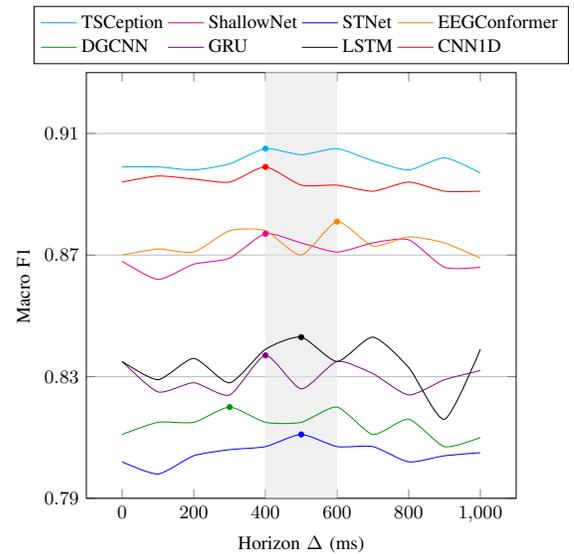
\begin{figure}[!t]
\centering
\resizebox{0.85\columnwidth}{!}{
\begin{tikzpicture}
\begin{axis}[
    width=0.99\columnwidth,
    height=0.48\textwidth,
    ymin=0.79,
    ymax=0.93,
    ytick={0.79, 0.83, 0.87, 0.91},
    ymajorgrids,
    ylabel={Macro F1},
    xlabel={Horizon $\Delta$ (ms)},
    legend columns=4,
    legend cell align={left},
    xtick={0,200,400,600,800,1000},
    legend style = {font=\small, at={(0.5,1.02)}, anchor=south},
    tick label style={font=\small},
    label style={font=\small},
    cycle list={
        {cyan, thick}, {magenta, thick}, {blue, thick}, {orange, thick},
        {green!60!black, thick}, {violet, thick}, {black, thick}, {red, thick}
    }
]
    \fill[black!10, opacity=0.6] (axis cs:400,0.79) rectangle (axis cs:600,0.93);
    \addplot[smooth, tension=0.4, cyan]          table[x=Delta,y=TSCeption] \resultsdata;
    \addplot[smooth, tension=0.4, magenta]       table[x=Delta,y=Shallow] \resultsdata;
    \addplot[smooth, tension=0.4, blue]          table[x=Delta,y=STNet] \resultsdata;
    \addplot[smooth, tension=0.4, orange]        table[x=Delta,y=EEGConformer] \resultsdata;
    \addplot[smooth, tension=0.4, green!60!black] table[x=Delta,y=DGCNN] \resultsdata;
    \addplot[smooth, tension=0.4, violet]        table[x=Delta,y=GRU] \resultsdata;
    \addplot[smooth, tension=0.4, black]         table[x=Delta,y=LSTM] \resultsdata;
    \addplot[smooth, tension=0.4, red]           table[x=Delta,y=CNN1D] \resultsdata;
    \begin{scope}[mark=*, mark size=1.2pt, only marks]
        \addplot[cyan] coordinates {(400, 0.905)};          
        \addplot[magenta] coordinates {(400, 0.877)};       
        \addplot[blue] coordinates {(500, 0.811)};          
        \addplot[orange] coordinates {(600, 0.881)};        
        \addplot[green!60!black] coordinates {(300, 0.820)}; 
        \addplot[violet] coordinates {(400, 0.837)};        
        \addplot[black] coordinates {(500, 0.843)};         
        \addplot[red] coordinates {(400, 0.899)};           
    \end{scope}
    \legend{TSCeption, ShallowNet, STNet, EEGConformer, DGCNN, GRU, LSTM, CNN1D};
\end{axis}
\end{tikzpicture}
}
\caption{Macro F1-scores for top 8 models. The grey shaded area (400--600 ms) indicates the neural preparatory window where optimal performance is observed.}
\label{fig:horizon_analysis}
\end{figure}

Early intention prediction was evaluated across horizons from 0 to 1000 ms ahead of action execution. \figurename~\ref{fig:horizon_analysis} shows Macro-F1 trajectories for the top-performing models.

Top-performing architectures exhibited strong temporal stability. TSCeption peaked at 400 ms ($\text{F1}=0.905$) and showed less than 1\% degradation at 1000 ms. ShallowConvNet maintained near-baseline performance across all horizons, while EEGConformer showed the largest improvement at future horizons, achieving a +1.26\% gain at 600 ms.

Across the top eight models, mean F1 remained stable between $0.852$ at 0 ms and $0.857$ at 400 ms, with the 400–600 ms range consistently yielding optimal performance. This window aligns with neuroscientific evidence that the late phase of the Bereitschaftspotential (readiness potential) reflects motor-intention formation several hundred milliseconds prior to movement execution~\cite{verleger2016time}. Notably, the majority of top-performing models exhibited less than 1.5\% degradation when predicting intentions 1s ahead, demonstrating suitability for anticipatory vehicle control.

\pgfplotstableread[row sep=\\,col sep=&]{
Delta & TSCeption & Shallow & STNet & EEGConformer & DGCNN & GRU & LSTM & CNN1D \\
0     & 0.899     & 0.868   & 0.802 & 0.870         & 0.811 & 0.835 & 0.835 & 0.894 \\
100   & 0.899     & 0.862   & 0.798 & 0.872         & 0.815 & 0.825 & 0.829 & 0.896 \\
200   & 0.898     & 0.867   & 0.804 & 0.871         & 0.815 & 0.828 & 0.836 & 0.895 \\
300   & 0.900     & 0.869   & 0.806 & 0.878         & 0.820 & 0.824 & 0.828 & 0.894 \\
400   & 0.905     & 0.877   & 0.807 & 0.878         & 0.815 & 0.837 & 0.839 & 0.899 \\
500   & 0.903     & 0.874   & 0.811 & 0.870         & 0.815 & 0.826 & 0.843 & 0.893 \\
600   & 0.905     & 0.871   & 0.807 & 0.881         & 0.820 & 0.835 & 0.835 & 0.893 \\
700   & 0.901     & 0.874   & 0.807 & 0.873         & 0.811 & 0.831 & 0.843 & 0.891 \\
800   & 0.898     & 0.875   & 0.802 & 0.876         & 0.816 & 0.824 & 0.833 & 0.894 \\
900   & 0.902     & 0.866   & 0.804 & 0.874         & 0.807 & 0.829 & 0.816 & 0.891 \\
1000  & 0.897     & 0.866   & 0.805 & 0.869         & 0.810 & 0.832 & 0.839 & 0.891 \\
}\resultsdata

\subsection{Limitations of the Proposed Work}
While the proposed framework demonstrates strong performance, several limitations should be noted. Generalisation across unseen subjects, routes, and vehicles was not explicitly evaluated. The dataset is limited to a small number of participants and a predefined route, which may introduce subject- and environment-specific biases. Future work will focus on increasing participants and conducting cross-subject and cross-environment evaluations to improve generalisability.
\section{Conclusion}
This study established a robust EEG-based framework for driver intention prediction, demonstrating that methodological choices in data preparation and model architecture are paramount. A 1.0-second windowing strategy with label-stratified temporal splitting significantly enhances reliability, while the systematic evaluation of twelve architectures revealed the superiority of domain-specific convolutional models, with TSCeption achieving the best performance ($\text{F1}=0.901$), followed by CNN1D ($\text{F1}=0.894$) and EEGConformer ($\text{F1}=0.874$). Minimal preprocessing consistently outperformed aggressive artefact-handling, indicating that raw EEG signals retain essential discriminative features in real-world conditions.

Results demonstrate stable long-horizon prediction, with optimal performance in the 400–600 ms window and less than 1.5\% degradation up to 1000 ms. Increasing spatial resolution to 16 channels further improved performance, supporting the feasibility of integrating EEG-based decoding into proactive safety systems.

However, generalisation across unseen subjects, routes, and vehicles remains untested, and the limited number of participants may introduce bias. Future work will address cross-subject evaluation, real-time ADAS integration, and testing under imbalanced data distributions.

\bibliographystyle{IEEEtran} 
\bibliography{bib/main}

\end{document}